\documentclass{article}
\usepackage{spconf,amsmath,epsfig}

\usepackage{amsmath}
\usepackage{amssymb}

\usepackage{algorithmic}

\usepackage{array}

\usepackage[english]{babel}
\usepackage{graphicx}
\usepackage{tablefootnote}

\usepackage{caption}
\usepackage{subcaption}
\usepackage{enumitem}
\usepackage{booktabs}

\usepackage[T1]{fontenc}
\usepackage{lmodern}

\usepackage{algorithm}
\usepackage{algorithmic}
\usepackage{comment}

\usepackage{multirow}

\usepackage{amsmath}

\usepackage{array}
\newcolumntype{P}[1]{>{\centering\arraybackslash}p{#1}}
\newcolumntype{M}[1]{>{\centering\arraybackslash}m{#1}}

\makeatletter
\DeclareRobustCommand{\rvdots}{%
  \vbox{
    \baselineskip4\p@\lineskiplimit\z@
    \kern-\p@
    \hbox{.}\hbox{.}\hbox{.}
  }}
\makeatother

\usepackage{tabularx,ragged2e}

\newcolumntype{Y}{>{\RaggedRight\arraybackslash\hspace{0pt}}X}
\newcolumntype{C}{>{\Centering\arraybackslash\hspace{0pt}}X}

\graphicspath{{./}{viz-images/hypermaps2/}}

\usepackage{tikz}
\usepackage{pgfplots}
\pgfplotsset{compat=newest}
\usetikzlibrary{shapes.geometric,arrows,fit,matrix,positioning}
\tikzset
{
    treenode/.style  = {},
    subtree/.style  = { anchor=north}
}

\newcommand\boldblue[1]{\textcolor{blue}{\textbf{#1}}}
\newcommand\boldred[1]{\textcolor{red}{\textbf{#1}}}

\title{An Inductive Mapping with Convolutional Representations for Human Settlement Detection: Preliminary Results}

\name{Dalton~Lunga, Dilip~Patlolla, Lexie~Yang, Jeanette~Weaver, and Budhendra~Bhadhuri}%
\address{Computational Sciences and Engineering Division \\ Oak Ridge National Laboratory}

\begin{document}
%
\maketitle
%

\begin{abstract}
We test this premise and explore representation spaces from a single deep convolutional network and their visualization to argue for a novel unified feature extraction framework. The objective is to utilize and re-purpose trained feature extractors without the need for network retraining on three remote sensing tasks i.e. superpixel mapping, pixel-level segmentation and semantic based image visualization. By leveraging the same convolutional feature extractors and viewing them as visual information extractors that encode different image representation spaces, we demonstrate a preliminary inductive transfer learning potential on multiscale experiments that incorporate edge-level details up to semantic-level information.
\end{abstract}
\section{Introduction}%
Even though the problem of image understanding has a long history in computer vision applications; recent breakthroughs in high performance computing and the availability of large overhead imagery are leading the cause for its surging appeal for disruptive remote sensing (RS) applications. Fueled by the success of methods including deep convolutional neural networks (CNN) in multimedia image classification, similar efforts are being sought for high resolution RS applications that not only achieve human level performance in terrestrial object classification but achieve semantic labeling and building extraction at large scale. However, due to its spatio-temporal nature, RS data offers unique challenges that necessitates the design of new deep neural architectures. For example,  by treating building extraction as an object localization challenge a deep learning implementation requires that both spatial and semantic image representations are combined towards training a convolutional classifier for accurate building-level detection\cite{Yuan2016}. Theoretical frameworks for understanding CNN architectures are yet to be fully explored. However, in the past few years, visualization technologies have emerged to close the gap by providing valuable insights on the information extraction stages of CNNs. Such crucial understanding is attributed to a variety of top performing deep CNNs\cite{Szegedy2015,Simonyan2014,Yu2016}. Training deep CNNs is computationally expensive and requires elusive knowledge on hyper-parameter tuning. Mild efforts are being directed to study the potential for multiple tasks to leverage on shared image CNN representations\cite{HariharanAGM14a}. 

It is with the above motivation that this paper seeks to explore the same representations towards image-level classification, pixel-level segmentation and semantic neighborhood mapping without the need for CNN re-training. An investigation is conducted by visually seeking to understand image representations via probing internal activation maps during the CNN forward pass process. Preliminary results shows that with limited labeling information, a unified representation learning of human settlement structures with overhead imagery has greater potential. Re-purposing of CNN feature extractors from course labels to fine-grained and semantic mapping tasks seeks to inform their wide applicability in RS. Using the visual understanding of CNNs, we highlight the following: (1) inductive transfer learning capabilities with a unified representation feature extractor toward multiple human settlement mapping tasks, (2) use semantic representational space to understand a collection of images, and (3) seeking maximal activation maps to obtain insights on per-class image characteristics to inform unique design of RS driven CNN architectures.

Image-level ground truth acquired from \textbf{0.5}-meter overhead imagery is used to train a CNN towards superpixel human settlement mapping. Training data consists of $40,000$ image patches, each of size $144\times 144$ pixels. From $4$-large tile scenes we crop out the training patches on a grid including $20,000$ image patches for the validation set. A second set of $20,000$ image patches is created from a different geographic location for generating a second semantic neighborhood mapping result. The CNN architecture consist of \textbf{7}-weight layers including \textbf{4}-convolutional (\textbf{conv}) layers, \textbf{2}-fully connected (\textbf{fcn}) layers, and \textbf{3}-maxpooling(\textbf{pool}) layers. \textbf{Pool} layers are configured after each \textbf{conv} layer. CNN odel parameters are obtained via a stochastic gradient descent (SGD) technique based on the back-propagation framework. The SGD learning rate is set to $0.00273$ via a full hyper-parameter gridsearch, while the activation is performed with ReLU, filter weights initialized from a normal distribution, and the batch size set to $150$. The trained CNN feature extractors are re-purposed to assess two different tasks i.e. (1) pixel-level mapping, (2) semantic and topological mapping. We present our early observations and highlight the essentials of probing CNN maps to inform multiscale tasking with a single representational learning learning framework.

\section{Related Work}%
Visual understanding of deep neural network architectures has enabled the capability to extract valuable insights on the internal transformations performed by the CNN filtering process in many computer vision tasks. Visual probing for insight extraction in CNNs can be traced back to the early work in \cite{Krizhevsky2012}, where a direct projection of first layer filters to the image space was performed to assess the learning capacity of the network. A technique to project intermediate and deeper CNN activation maps was demonstrated in \cite{Zeiler2013} via a deconvolution process to yield reconstructed image encodings that were interpreted via the discriminant strong information from the input image pixels. In \cite{Simonyan2014}, image patches that maximize selected neuron activation maps were sought by performing a gradient ascent in the image space with the goal of studying their pixel level characteristics. In \cite{Mahendran2014}, the authors extend the method towards seeking input images that trigger similar neural stimulation for a given layer. The work in \cite{Yu2016} sought to visualize in a more comprehensive manner the representation spaces constructed by all filters of a layer.%

We draw lessons from this body of literature to seek insights on the representation spaces constructed by CNNs  with \textbf{0.5}-meter single band remote sensing imagery. Our main objective being to leverage the representation spaces obtained with image-level ground truth toward assessing fine-grained settlement detection and mapping of images onto a semantic topological geometry.%

\section{Inductive Mapping with Overhead Images}
\begin{figure}[!ht]
  \centering
    \includegraphics[width=8cm,height=4.6cm]{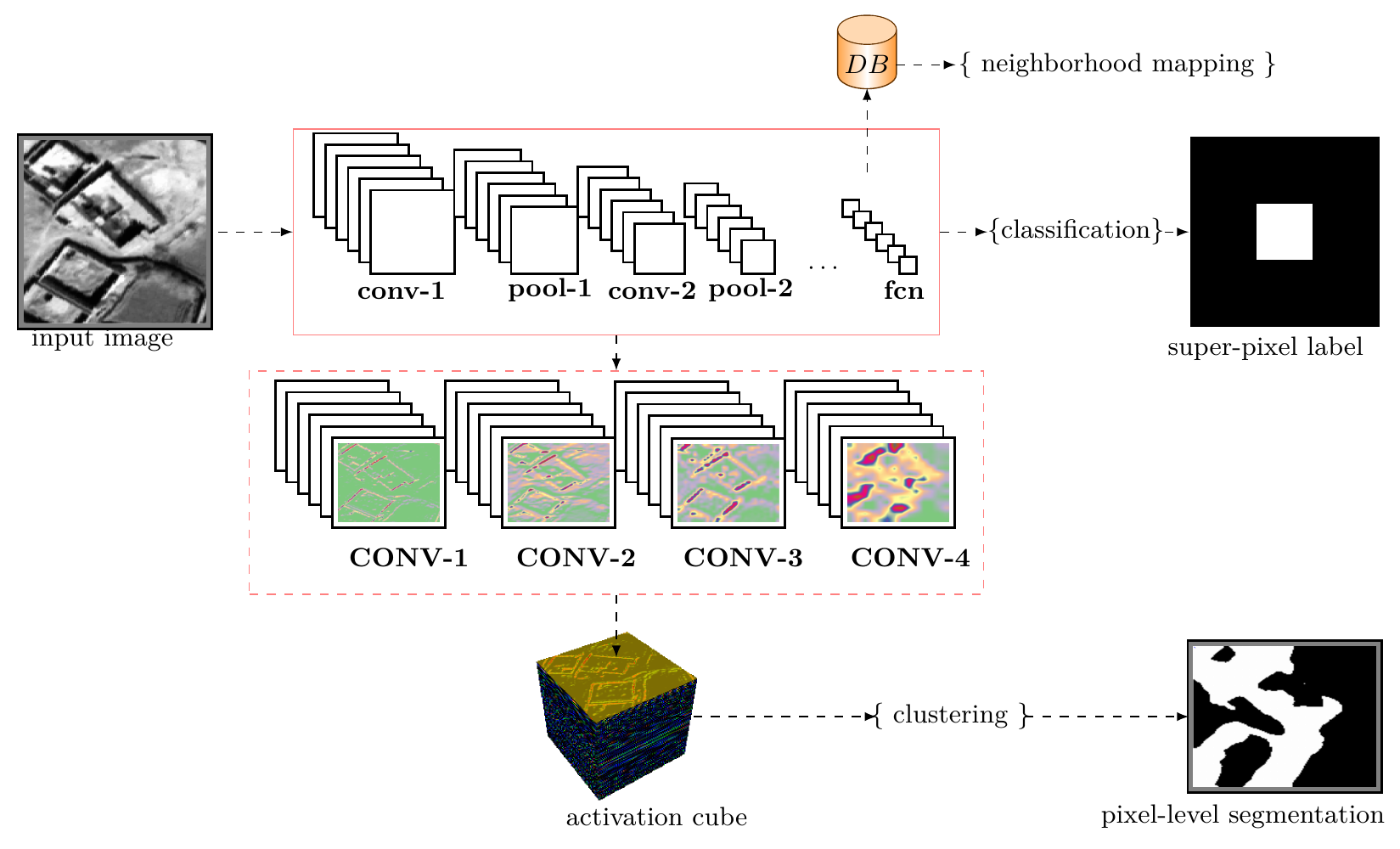}
\caption{A unified representational learning with CNN and inductive settlement extraction framework.}
\label{fig:representation-learning-framework}
\end{figure}

Remote sensing overhead image data reside on a spatio-temporal grid and this is in contrast to independent and identically distributed sample-based methodologies widely adopted in traditional machine learning. Human settlement mapping is a typical challenge that requires that an image representation should take the grid nature of the data into account. CNNs have proven to be applicable in leveraging the spatial image grid\cite{Krizhevsky2012,Paisitkriangkrai2016}. Given the range and complex nature of human settlement understanding with overhead imagery, the inductive transfer learning approach is to exploit visualization driven insights and demonstrate a single CNN framework on: (one) superpixel classification of homogeneous regions into single categories, (two) pixel-level classification to seek fine-grained detection of settlement structure boundaries and (three) using image-patches to compute for semantic level neighbourhood mapping. Figure~\ref{fig:representation-learning-framework} illustrate the generalized conceptual architecture for the envisioned unified representational learning framework with overhead imagery.

\subsection{Superpixel based settlement detection}
Homogeneous region classification is emerging in popularity with remote sensing imagery. Clear-cut boundaries are sought to distinguish classes (e.g. urban vs forest)\cite{Sethi2015}.
We illustrate the potential applicability of CNN features on a $16\times 16$ superpixel block region for human settlement mapping. We posit that using a large volume of local homogeneous patch level ground truth for training the representation learning framework, could offer greater potential to stimulate automatic learning of coherent local structures that are characteristic and common with overhead imagery containing human settlements. The hierarchical and deeper feature abstraction by CNNs could efficiently generate scale invariant representations that are favourable for seeking homogeneous regions. Figure~\ref{fig:mappings}(a) and (b) illustrate the superpixel based mapping generated with a softmax classification on the \textbf{fcn} features of the CNN.
\subsection{Hypercolumns for settlement detection}
\begin{figure}[ht] 
    \begin{subfigure}[b]{0.5\linewidth}
            \centering
            \includegraphics[width=4cm,height=3.8cm]{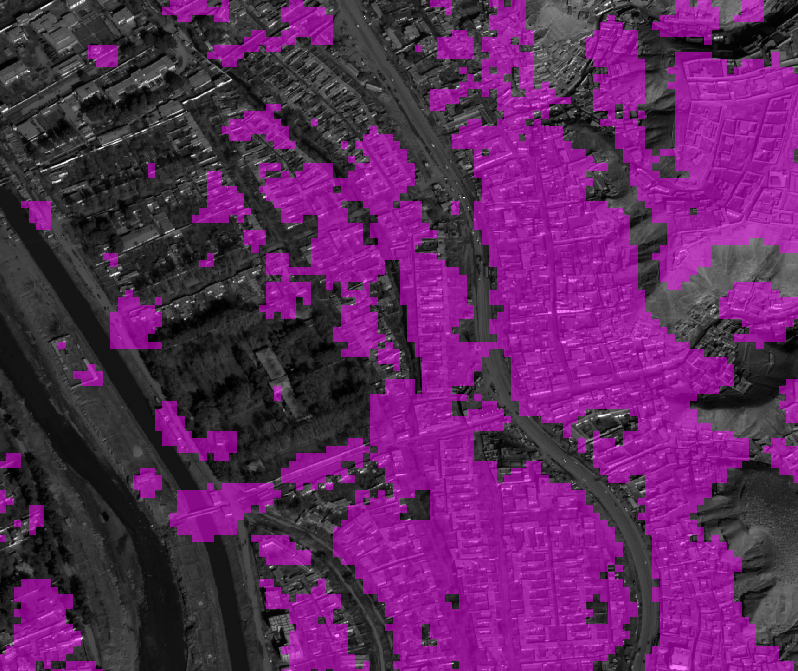} 
            \caption{ {\small dense settlements} }
            \label{fig7:a} 
            \vspace{2ex}
    \end{subfigure}
    \begin{subfigure}[b]{0.5\linewidth}
            \centering
            \includegraphics[width=4cm,height=3.8cm]{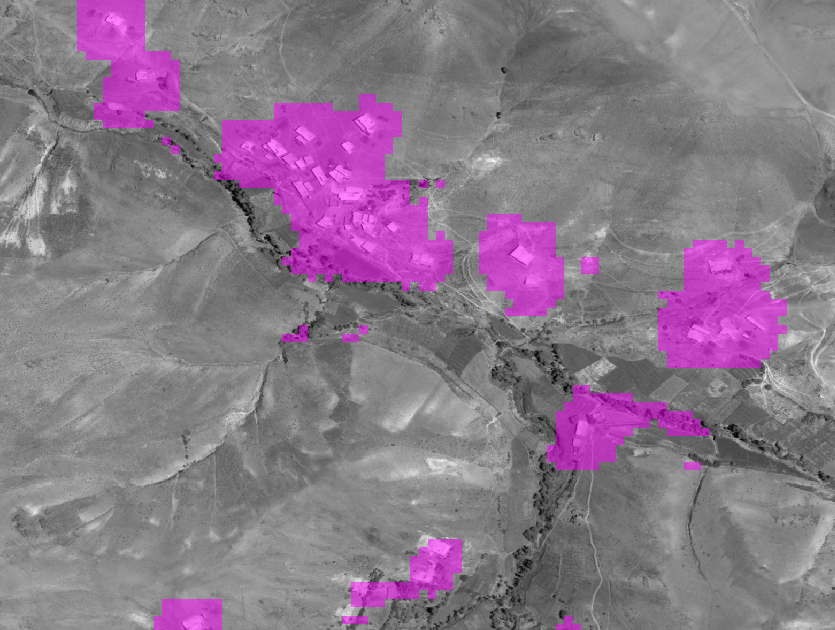} 
            \caption{{\small sparse settlements}}
            \label{fig7:b} 
            \vspace{2ex}
    \end{subfigure} 
    \begin{subfigure}[b]{0.5\linewidth}
            \centering
            \includegraphics[width=4cm,height=3.8cm]{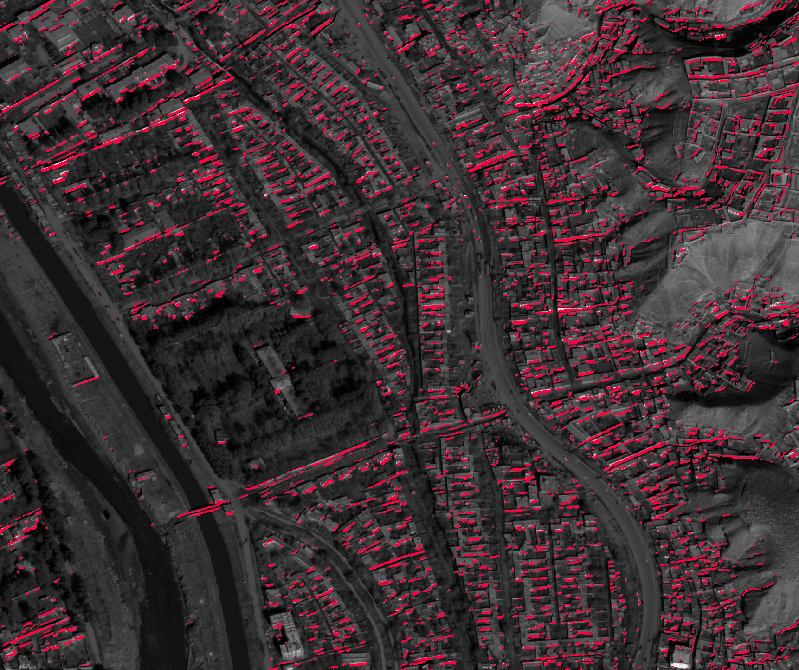} 
            \caption{{\small dense settlements}}
            \label{fig7:c} 
    \end{subfigure}
    \begin{subfigure}[b]{0.5\linewidth}
            \centering
            \includegraphics[width=4cm,height=3.8cm]{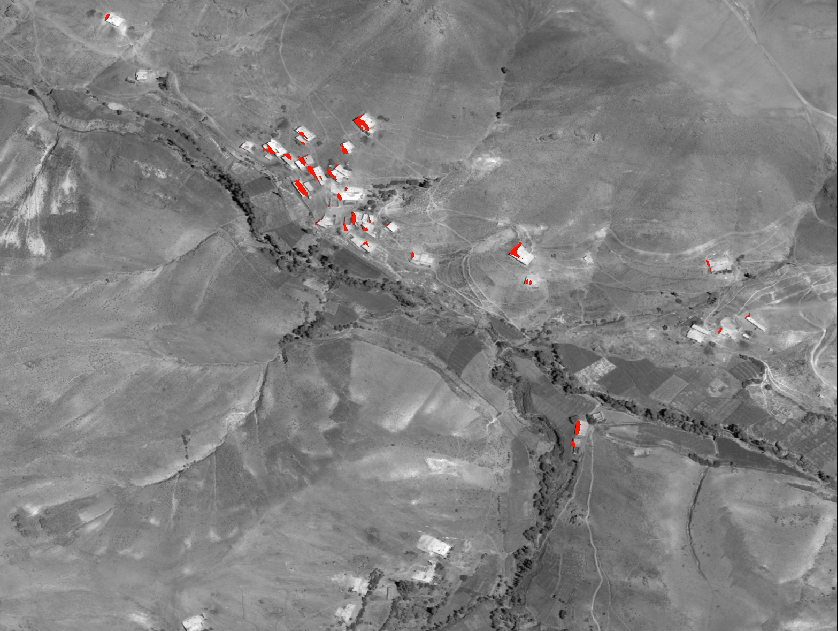} 
            \caption{{\small sparse settlements}}
            \label{fig7:d} 
    \end{subfigure} 
    \caption{Illustrating superpixel mapping(top row) and pixel-level terrain segmentation(bottom row) on \textbf{0.5}-meter $1728\times 1728$ pixel aerial imagery.}
    \label{fig:mappings} 
\end{figure}

It is inarguable that, at large, recognition algorithms based on CNNs typically use the output of the \textbf{fcn} layer as a feature representation. However, as shown by the example in Figure~\ref{fig:representation-learning-framework}, the information  from the top \textbf{conv} layers and the \textbf{fcn} layers appears to be too coarse spatially to enable precise pixel level settlement segmentation. As first demonstrated in \cite{HariharanAGM14a} and also reflected in Figure~\ref{fig:representation-learning-framework}, lower \textbf{conv} layers do retain edge detection information that may be precise in detecting settlement boundaries albeit not capture higher level semantic details to describe the settlement as a whole. Using the approach of \cite{HariharanAGM14a}, we define the hypercolumn at a pixel as the vector of activations of all CNN layers above that pixel. Using the hypercolumns as pixel descriptors, we leverage on image level trained CNN feature extractor to generate the fine-grained mapping shown in Figure~\ref{fig:mappings}(c) and (d). The algorithmic implementation utilizes a mini-batch K-means clustering algorithm to generate the pixel-level segmentation results. Figure~\ref{fig:mappings}(c) shows the pixel-segmentation results detecting additional settlements that are appear to be omissions in the top-left corner of the superpixel mapping in (a).
\subsection{Semantic and topological mapping}
Semantic image visualization using very high resolution remote sensing imagery has emerged as another challenging application in the past decade. The most recent attempt at this challenge presented a semi-supervised learning framework employing the notion of superpixel tessellation representations of imagery\cite{Sethi2015}. The image representation utilizes homogeneous and irregularly shaped regions and relies on hand designed features based on intensity histograms, geometry, corner and superpixel density and scale of tessellation. Of relation, we demonstrate the potential of leveraging top layer \textbf{fcn} representations, toward a semantic image representation visualization with thousands of image patches cropped from large scenes. Although the intermediate stages of the CNN could offer representations useful in related mapping tasks, the result in Figure~\ref{fig:representation-spaces} reveals a more homogeneous image content mapping based on top level features. By applying clustering methods in the projected space course level neighborhood segmentation map can potentially be generated.

\section{Image Information Extraction}
\begin{figure}[!ht]
\centering
\includegraphics[width=.48\textwidth]{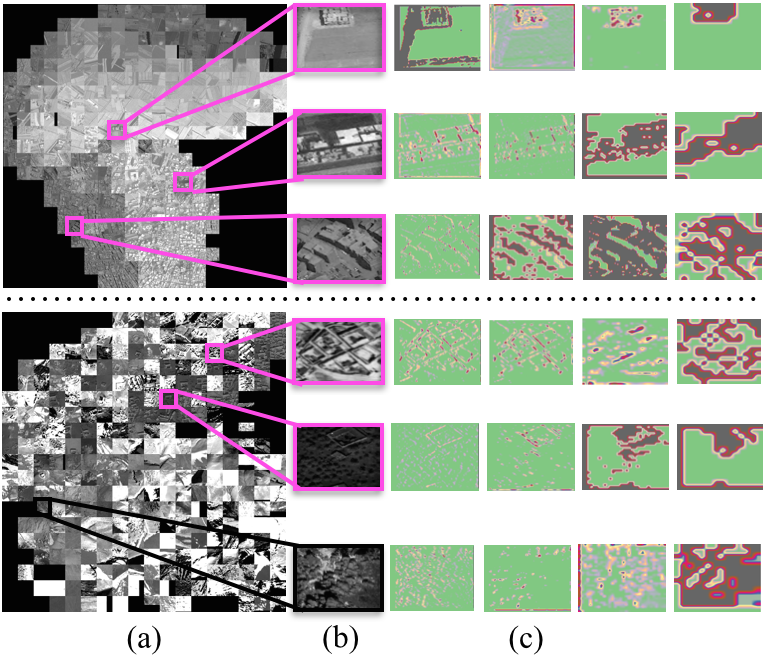}
\caption{Topological and semantic mapping of image information for Egypt (top row) and Afghanistan (bottom row). Column (a) shows the semantic neighborhood embedding plane using \textbf{fcn} representations. Column (b) are example patches from (a). Column (c) shows reconstructed maximally activated \textbf{conv} feature extractors for example patches in (b).}%
\label{fig:representation-spaces}
\end{figure}

Each stage of the \textbf{conv} filtering process retains class related discriminatory information with the top \textbf{conv} and \textbf{fcn} retaining object specific semantic information. Figure~\ref{fig:representation-spaces} shows a large scale semantic patch representation obtained with same CNN model for two different geographic locations. The result in (a) is a manifold learning projection\cite{Lunga2014,Maaten2008} of the \textbf{fcn} layer features. The projection shows patches close(similar in content) in the \textbf{fcn} representation space embedded close in the two-dimensional topological space. The gradual information extraction by CNNs can be visualized to gain useful insights for image representation understanding.  As shown in  (c), highest activation feature maps can be visualized on each layer to reveal discriminatory and useful information for relevant multiscale analysis. Edge-level and semantic level discriminatory information is consistently extracted across the six example patches in (b) containing {\em houses, rocks, roads} and {\em trees}.
\section{Conclusion}
A preliminary investigation is conducted  to exploit image-level ground truth towards a single CNN representational model in multiscale  human settlement understanding with \textbf{0.5}-meter resolution overhead imagery. We visually contrast our early results for superpixel mapping and the hypercolumn induced pixel-level segmentation. The investigation demonstrates a surprising potential to obtain reasonable fine-grained human settlement mapping from image-level ground truth. This outcome offers further grounds to expand our efforts to seek multiple objective functional forms towards RS mapping with a unified representation learning framework. In addition, the discriminatory capable \textbf{fcn} features when transformed via manifold projections pointed to another direction for seeking image segmentation at semantics level - a very promising avenue for seeking region-based RS semantic neighborhood mapping.

\section*{Acknowledgement}
This manuscript has been authored by UT-Battelle, LLC under Contract No. DE-AC05-00OR22725 with the U.S. Department of Energy. The United States Government retains and the publisher, by accepting the article for publication, acknowledges that the United States Government retains a non-exclusive, paid-up, irrevocable, world-wide license to publish or reproduce the published form of this manuscript, or allow others to do so, for United States Government purposes.

The authors would like to thank Dr. Jiangye Yuan and Dr Mark Coletti for discussions on related topics.

\bibliographystyle{IEEEbib}
\bibliography{overheard-dcnn}

\end{document}